\pdfoutput=1
\documentclass[10pt,twocolumn,letterpaper]{article}

\usepackage[pagenumbers]{cvpr}

\usepackage[T1]{fontenc}
\usepackage{microtype}

\newcommand{\Ls}{L^{\ast}}
\newcommand{\abs}{a^{\ast}b^{\ast}}
\newcommand{\dawn}{\textup{\textsc{Dawn}}}
\newcommand{\dawnf}{\textup{\textsc{Dawn}}-Flow}
\newcommand{\Ltok}{\mathcal{L}_{\text{tok}}}
\newcommand{\Lrender}{\mathcal{L}_{\text{render}}}
\newcommand{\Lalign}{\mathcal{L}_{\text{align}}}
\newcommand{\Lcolor}{\mathcal{L}_{\text{colour}}}

\definecolor{cvprblue}{rgb}{0.21,0.49,0.74}
\usepackage[pagebackref,breaklinks,colorlinks,allcolors=cvprblue]{hyperref}
\hypersetup{pdftitle={DAWN: Regression and Flow Matching on LoRA-Tuned DINOv3 for Thermal-to-Visible Drone Imagery}, pdfauthor={Chaitanya Dumbre, Rachit Udaseen, Trupti Baraskar}}

\def\confName{CVPR}
\def\confYear{2027}

\title{\dawn: Regression and Flow Matching on LoRA-Tuned DINOv3\\for Thermal-to-Visible Drone Imagery}

\author{Chaitanya Dumbre \qquad Rachit Udaseen \qquad Trupti Baraskar\\
School of Computer Science and Engineering, MIT World Peace University, Pune, India\\
{\tt\small chaidumb04@gmail.com, rachitudaseen@gmail.com, trupti.baraskar@mitwpu.edu.in}\\[0.6em]
{\small Code: \href{https://github.com/NotArsenic/Dawn-Paper-Release}{\texttt{github.com/NotArsenic/Dawn-Paper-Release}}}\\
{\small Weights: \href{https://huggingface.co/Arsenicc/Dawn-Paper-Release}{\texttt{huggingface.co/Arsenicc/Dawn-Paper-Release}}}}

\begin{document}
\maketitle

\begin{abstract}
In daytime thermal-to-visible drone imagery, lightness within a local region is largely a function of
thermal intensity, whereas colour is ambiguous. Based on this structure, \dawn{} represents lightness as an
affine bilateral grid over thermal intensity alongside a low-resolution chroma map, and predicts both
in a single pass of a 7.8M-parameter transformer of six DiT blocks conditioned on a LoRA-tuned DINOv3 encoder. On 781 held-out
box-aligned DroneVehicle test pairs, fitting the lightness grid directly to the photographs reaches
24.45\,dB $\Ls$ PSNR, an approximate ceiling for this representation. Training the same network by direct
regression (\dawn) and by conditional flow matching (\dawnf), and comparing both with pix2pix on the same
pairs, separates accuracy, realism and uncertainty. Single-pass regression (\dawn) is the most accurate,
reaching 20.48\,dB and a CIEDE2000 of 7.84 (mean of three seeds) in 11.02\,ms per frame on an RTX 5090. On
the same pairs, pix2pix produces sharper, more colourful images at the cost of higher distortion (18.56\,dB,
10.92 CIEDE2000). Trained with conditional flow matching (\dawnf, 20.01\,dB with two Euler steps), the same
architecture yields per-pixel sample-spread uncertainty whose mean per image ranks test frames by $\Ls$ error
(Spearman $\rho=0.77$) but underestimates it about fivefold. Replacing the tuned DINOv3 encoder with a small
CNN costs 0.61\,dB and keeping DINOv3 frozen costs 0.24\,dB, while its gated attention prior, the training
schedule, the token-loss weighting and guided chroma upsampling bring no measurable gain.
\end{abstract}

\begin{figure}[t]
\centering
\footnotesize
\makebox[0.2\linewidth]{Thermal}\makebox[0.2\linewidth]{\dawn}%
\makebox[0.2\linewidth]{\dawnf}\makebox[0.2\linewidth]{Photo}%
\makebox[0.2\linewidth]{$\sigma_{ab}$}\\[1pt]
\includegraphics[width=\linewidth]{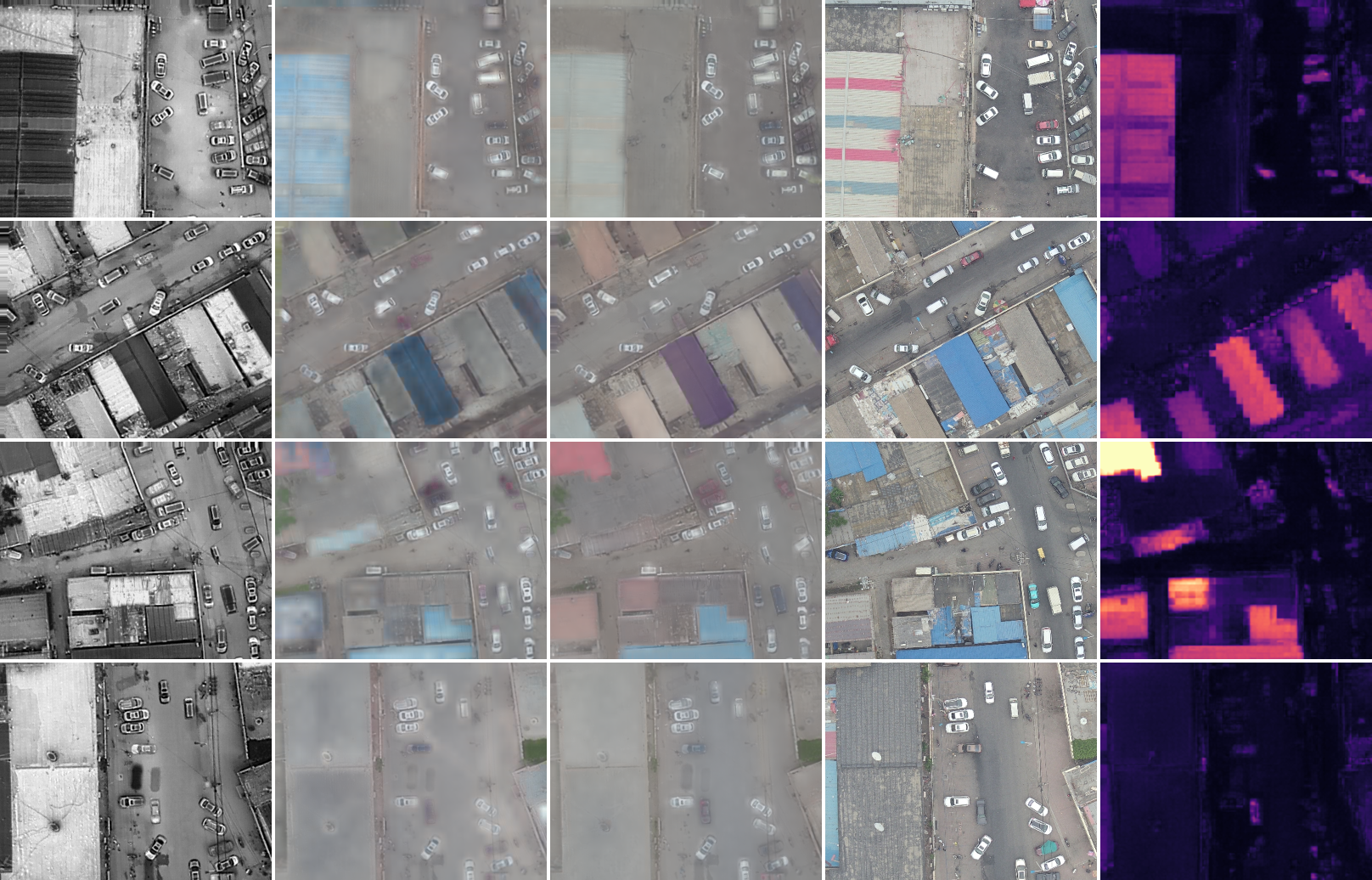}
\caption{\dawn{} on four test frames (bottom row fallback-aligned). From left to right: thermal input,
\dawn, one of eight samples of \dawnf, the photograph, and the spread of the eight \dawnf{} samples in
chroma ($\sigma_{ab}$, on a fixed scale, brighter is larger). Where the colour is ambiguous, as on the roofs,
the spread is large and the two models can choose different colours. Over the test set, both models
are less colourful than the photographs (Tab.~\ref{tab:main}). Images from DroneVehicle~\cite{sun2022dronevehicle}.}
\label{fig:teaser}
\end{figure}

\begin{figure*}[t]
\centering
\includegraphics[width=\textwidth]{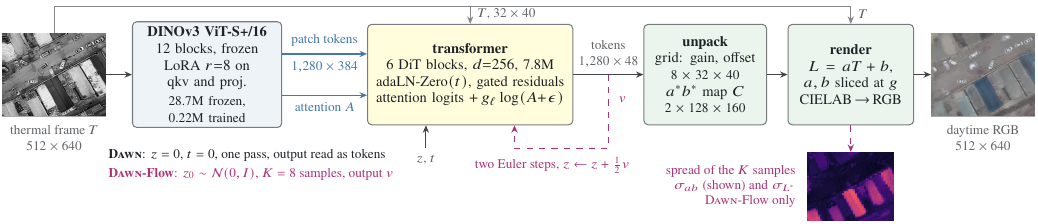}
\caption{Overview. A frozen DINOv3 ViT-S+/16 with rank-8 LoRA on the qkv and output projections of its
12 blocks encodes the thermal frame. Projections of its patch tokens and of a $32\times40$ downsampled thermal
frame are added to the input tokens of a six-block DiT, and its last-layer attention enters every block as a
logit bias behind a gate initialised at zero (Eq.~\eqref{eq:gate}). Time $t$ enters through adaLN-Zero,
gating every residual branch so each block starts as the identity. \dawn{} runs the transformer once with
$z=0$ and $t=0$ and reads its output directly as tokens. \dawnf{} reads the output as a velocity $v$ and takes
two Euler steps from noise for each of $K$ samples, whose spread gives the uncertainty maps
(Eq.~\eqref{eq:spread}, here $\sigma_{ab}$). The tokens unpack into the brightness grid and the $\abs$ map
$C$, rendered in CIELAB (Eqs.~\eqref{eq:grid} and~\eqref{eq:render}). Thumbnails show the second row of
Fig.~\ref{fig:teaser}. Images from DroneVehicle~\cite{sun2022dronevehicle}.}
\label{fig:arch}
\end{figure*}

\section{Introduction}
\label{sec:intro}

Thermal images are hard to interpret for untrained operators~\cite{berg2018generating}, and their
low contrast and missing colour limit the use of RGB-trained vision models~\cite{luo2024foalgan}.
Translating a thermal frame into the RGB photograph that a colour camera would have recorded addresses
both problems, but the task is ill-posed. Lane markings, text or the paint of a car leave no thermal
signature and can only be inferred from context. We accept such invented detail as long as the layout
follows the thermal input, and ideally the model should also tell where it is guessing.

Our design rests on two observations. First, within a local region the brightness of a daylight
image is largely a function of thermal intensity. Asphalt, roofs and vegetation each relate
temperature to brightness in a fairly stable way that changes across the frame. A spatially varying
affine curve then explains most of the lightness, and how much it explains can be measured before
training. Second, colour is ambiguous. Nothing in a thermal image distinguishes a red car from a blue
one. A regression model averages such alternatives toward grey, whereas a generative model can commit
to one of them and, by sampling, show where it is unsure.

\dawn{} predicts the coefficients of a brightness grid and a $128\times160$ chroma map
instead of pixels, with a small transformer conditioned on a LoRA-adapted~\cite{hu2022lora} DINOv3
encoder~\cite{simeoni2025dinov3} (Fig.~\ref{fig:arch}). To test the second observation, we train the
same network twice, as a direct regressor (\dawn) and as a rectified flow~\cite{liu2023rectified}
(\dawnf). The regressor is the more accurate of the two (20.48 vs.\ 20.01\,dB $\Ls$ PSNR over three
seeds) and needs a single network pass. The spread of the flow model's samples follows the error of a frame but
underestimates it, and more sampling steps give more colour while accuracy drops further. In this
setting the generative formulation pays for its uncertainty maps with accuracy. This is why we present
the regressor as our model and the flow model as its uncertainty variant.

All reported numbers are computed from stored training logs and evaluation outputs. We reran
the data pipeline from the raw DroneVehicle archives, the oracle on the validation pairs and the
evaluation of \dawnf{} (seed 42) on a second machine (RTX 5060 Ti), and repeated that evaluation
on a second RTX 5090 and on a CPU. The reruns reproduce Tab.~\ref{tab:funnel}, the validation
columns of Tab.~\ref{tab:oracle} and the seed-42 \dawnf{} test scores, with per-frame metric
differences of at most $2\times10^{-4}$. Our contributions are summarised below.
\begin{itemize}\setlength{\itemsep}{1.5pt}\setlength{\parskip}{0pt}\setlength{\topsep}{3pt}
\item A thermal-to-visible representation whose lightness is an affine bilateral grid of thermal intensity and whose colour is a coarse chroma map, with an attainable accuracy that can be estimated before training.
\item A controlled comparison of the same network trained as a direct regressor (\dawn) and as a rectified flow (\dawnf), benchmarked with pix2pix~\cite{isola2017pix2pix} and two global baselines to separate accuracy, realism and uncertainty.
\item Ablations showing that LoRA-tuned DINOv3 features account for part of the accuracy, whereas DINOv3's gated attention prior, the training schedule, the loss weighting and the chroma upsampling make no measurable difference.
\item A cross-modal alignment pipeline yielding 5,385 daytime DroneVehicle pairs, together with an analysis of the flow model's uncertainty, colour collapse, runtime and gate dynamics.
\end{itemize}

\section{Related Work}
\label{sec:related}

\paragraph{Thermal-to-visible translation.} CNNs have been used to colourise near-infrared
images~\cite{limmer2016infrared}, where high-frequency detail can still be transferred from the input.
For thermal infrared, Berg et al.~\cite{berg2018generating} note that greyscale colourisation cannot be
applied directly because visible luminance is unknown as well, and train a CNN to predict CIELAB images. Later work uses conditional GANs~\cite{kuang2020thermal},
translates nighttime thermal images into daytime colour images~\cite{luo2024foalgan}, or applies
diffusion models to faces~\cite{nair2023t2v}. Like general image-to-image translation, whether adversarial~\cite{isola2017pix2pix,zhu2017cyclegan} or based on diffusion~\cite{ho2020ddpm} as in Palette~\cite{saharia2022palette}, these networks predict pixels, whereas \dawn{} predicts the parameters of a structured image model.

\paragraph{Colourisation, flows and bilateral grids.} Automatic
colourisation~\cite{zhang2016colorful,larsson2016learning} faces the same ambiguity, and we pool the
soft-encoded pixel targets of Zhang et al.~\cite{zhang2016colorful} into patch histograms as an auxiliary loss. Flow matching, rectified flow and stochastic interpolants~\cite{lipman2023flow,liu2023rectified,albergo2023interpolants} learn a velocity field that transports noise to data, and linear interpolation~\cite{lipman2023flow,liu2023rectified} gives low-curvature paths suited to few-step sampling. Our transformer is a small DiT~\cite{peebles2023dit}. The bilateral grid~\cite{chen2007bilateral} enables fast edge-aware processing. HDRNet~\cite{gharbi2017hdrnet} predicts local affine transforms in such a grid with a CNN, and bilateral guided upsampling (BGU)~\cite{chen2016bgu} fits them to upsample an operator. We use such a grid across modalities, as affine models of thermal intensity whose coefficients come from DINOv3 features.

\paragraph{Foundation features, gating and uncertainty.} We adapt DINOv3 ViT-S+/16~\cite{simeoni2025dinov3}, a
ViT~\cite{dosovitskiy2021vit} with register tokens~\cite{darcet2024registers}, rotary position embeddings~\cite{su2024roformer} and SwiGLU feed-forward layers~\cite{shazeer2020glu}, using
LoRA~\cite{hu2022lora}. ReZero~\cite{bachlechner2021rezero}, Flamingo~\cite{alayrac2022flamingo} and DiT's
adaLN-Zero~\cite{peebles2023dit} place zero-initialised gates on residual branches to keep each block an identity map at the start of training. Disagreement among ensemble members~\cite{lakshminarayanan2017ensembles},
stochastic forward passes~\cite{gal2016dropout,kendall2017uncertainties} or repeated samples of a conditional diffusion model~\cite{whang2022deblurring} is a common uncertainty proxy, and pixel-wise uncertainty in diffusion models has also been estimated via Bayesian inference~\cite{kou2024bayesdiff}.

\section{Method}
\label{sec:method}

Fig.~\ref{fig:arch} gives an overview. The input is a thermal frame $T\in[0,1]^{H\times W}$,
$H\times W=512\times640$ (8-bit codes divided by 255), and the target $y$ is the paired photograph in
linear RGB. Grid cells, DINOv3 patches and the transformer's tokens share one $32\times40$ grid of
$16\times16$-pixel cells ($N=1{,}280$).

\paragraph{Lightness as a bilateral grid.} Let $g$ be $T$ smoothed with a $5\times5$ box filter. The
grid stores a gain $a_n$ and an offset $b_n$ at each node $n=(k,i,j)$, where $k\in\{0,\dots,7\}$ indexes
eight thermal-intensity bins and $(i,j)$ the spatial cell. Each pixel $p$ interpolates its eight
neighbouring nodes $\mathcal N(p)$ with trilinear weights $w_n(p)$, placed at cell centres in space
and at bin centres along $g(p)$. The lightness $L=\Ls/100$ becomes an affine function of $T$ whose
coefficients vary smoothly in space and intensity:
\begin{equation}
\hat L(p)=\operatorname{clamp}_{[0,1]}\Big(\textstyle\sum_{n\in\mathcal N(p)} w_n(p)\,\big[a_n T(p)+b_n\big]\Big).
\label{eq:grid}
\end{equation}
Target grids are fitted per image, without gradients, by weighted ridge regression~\cite{hoerl1970ridge}, with each node
regularised toward the fit of its cell and each cell toward an image-level fit, followed by one
refinement pass on the residual.

\paragraph{Chroma.} The model also predicts an $\abs$ map $C$ at $128\times160$, \ie, a $4\times4$
block per token, whose target is the masked mean of the photograph's $\abs$ over each $4\times4$-pixel
block. It is upsampled bilinearly and combined with the lightness in CIELAB~\cite{robertson1977cielab},
\begin{equation}
\hat y=\operatorname{clip}_{[0,1]}\big(\operatorname{Lab}^{-1}(100\,\hat L,\ \operatorname{Up}(C))\big).
\label{eq:render}
\end{equation}

\paragraph{Two ways to predict the tokens.} Each token holds the 16 grid coefficients of its cell (a
gain and an offset per bin) and the 32 $\abs$ values of its chroma block, $x_n\in\mathbb R^{48}$.
Tokens are standardised per grid coefficient, intensity bin and chroma channel using statistics from
2,000 training pairs. A transformer~\cite{vaswani2017attention} $f_\theta(z,t,c)$ maps a token array $z$, a time $t$ and the
conditioning $c$ to an array of the same shape. With $z_1$ the target tokens and
$z_t=(1-t)\,z_0+t\,z_1$, both of our models minimise
\begin{equation}
\Ltok=\sum_{s\in\{\text{bri},\text{col}\}} \tfrac{D_s}{D}\,
\mathbb E\big[\,\overline{\omega_s\odot(f_\theta(z_t,t,c)-(z_1-z_0))^{2}}\,\big],
\label{eq:flow}
\end{equation}
where $s$ ranges over the brightness and chroma parts of the token, $D_{\text{bri}}=16$,
$D_{\text{col}}=32$, $D=48$, and the overline denotes the mean over a part. The weights $\omega$
give each brightness node its trilinearly weighted count of valid pixels and each chroma value the
unmasked fraction of its pixel, normalised to mean one per part and sample. \dawn{} fixes $t=0$ and $z_0=0$, so that
$f_\theta$ regresses the target tokens directly, and predicts $\hat z_1=f_\theta(0,0,c)$ in a single
pass. \dawnf{} draws $z_0\sim\mathcal N(0,I)$ and $t\sim\mathcal U[0,1]$, which makes $f_\theta$ the
velocity field of a conditional rectified flow~\cite{liu2023rectified} (equivalently, linear conditional flow matching~\cite{lipman2023flow}), trained without reflow. At inference it
takes two Euler steps of size $\tfrac12$ from noise. In both cases, predicted tokens are rescaled by
the training-set mean and standard deviation, unpacked into the lightness grid and chroma map $C$, and
rendered via Eqs.~\eqref{eq:grid} and~\eqref{eq:render}.

\paragraph{Conditioning.} The encoder is a frozen DINOv3 ViT-S+/16~\cite{simeoni2025dinov3} (28.7M parameters) applied to the
thermal frame replicated to three channels. Rank-8 LoRA adapters~\cite{hu2022lora} ($\alpha=16$) on the qkv and output
projections of all blocks add 0.22M trainable parameters in 24 modules, with zero-initialised up-projections
so training starts from the pretrained encoder. It provides 1,280 patch tokens and the
patch-to-patch attention $A$ of its last layer, averaged over heads, restricted to patch tokens,
renormalised per row and detached. The transformer has six DiT blocks~\cite{peebles2023dit} of
width 256 with four heads (7.82M parameters). Each token's input embedding adds a learned positional
embedding to linear projections of $z_n$, the matching DINOv3 patch token and the thermal frame
area-averaged to $32\times40$. The time $t$ enters every block through adaLN-Zero, which makes
each block the identity at initialisation, and the output projection is also zero-initialised. Every block
$\ell$ adds DINOv3's log attention to its own attention logits, scaled by a learned scalar $g_\ell$
initialised to zero:
\begin{equation}
\operatorname{Attn}_\ell(Q,K,V)=\operatorname{softmax}\!\Big(\tfrac{QK^{\top}}{\sqrt d}+g_\ell\log(A+\epsilon)\Big)V,
\label{eq:gate}
\end{equation}
with $\epsilon=10^{-6}$. Since the bias is a logarithm, $g_\ell$ acts as an exponent on $A$. At
$g_\ell=0$ the block ignores the prior, and at $g_\ell=1$ the two distributions are multiplied.
Detaching $A$ blocks gradients from flowing back into the encoder through the attention bias. Each gate
$g_\ell$ still receives gradients from the attention output and is excluded from weight decay.

\paragraph{Training objective.} We minimise
\begin{equation}
\mathcal L=\Ltok+0.1\,\Lrender+0.5\,\Lalign+0.2\,\Lcolor,
\label{eq:objective}
\end{equation}
with all terms masked as described in Sec.~\ref{sec:data}. Let $\hat Y$ and $Y$ denote the $256\times320$
render of the one-step estimate $\hat z_1=z_t+(1-t)\,f_\theta$ (for \dawn, its direct prediction) and the
downsampled photograph, $m_p$ the pixel mask and $e_p$ the absolute difference $|\hat Y_p-Y_p|$, averaged
over the three CIELAB channels and divided by 100. The rendering loss is
\begin{equation}
\Lrender=\mathbb E\Big[w_{\text{pair}}\,\frac{\sum_p m_p e_p}{\sum_p m_p}\Big],
\label{eq:lrender}
\end{equation}
with pair weight $w_{\text{pair}}=0.5$ on fallback-aligned pairs and $1$ otherwise. Feature alignment
compares the adapted encoder's thermal patch tokens $u_n$ with the frozen encoder's tokens $u^{\text{rgb}}_n$
on the paired RGB image, weighting each patch by its unmasked fraction $\alpha_n$ and summing over all
patches in the batch,
\begin{equation}
\Lalign=\frac{\sum_n \alpha_n\big(1-\cos(u_n,u^{\text{rgb}}_n)\big)}{\sum_n \alpha_n}.
\label{eq:lalign}
\end{equation}
For $\Lcolor$, each valid pixel is soft-encoded onto its five nearest neighbours ($\sigma=5$) as in
Zhang et al.~\cite{zhang2016colorful}, over the 261 cells of a 10-unit $\abs$ grid occupied by sRGB,
giving a per-patch target histogram $q_n$. A linear head on layer-normalised $u_n$ predicts class
probabilities $\hat p_n$, trained with the cross-entropy
\begin{equation}
\Lcolor=-\frac{\sum_{n\in\mathcal P}\alpha_n\,q_n^{\top}\log\hat p_n}{\sum_{n\in\mathcal P}\alpha_n},
\label{eq:lcolor}
\end{equation}
where $\mathcal P=\{n:\alpha_n\ge 0.5\}$ leaves out patches less than half inside the mask. Unlike
Zhang et al.~\cite{zhang2016colorful}, we do not rebalance the classes. $\Lalign$ and $\Lcolor$ train
LoRA and the head (0.10M parameters, discarded at inference) and nothing else.

\paragraph{Detach-then-ramp schedule.} For the first 3,000 steps we stop gradients from the transformer
into the patch tokens, training LoRA with $\Lalign$ and $\Lcolor$ alone. The gradient of $\Ltok$ and
$\Lrender$ into the encoder is then scaled linearly from zero to one over 500 steps, without changing
the forward pass.

\paragraph{Uncertainty.} For \dawnf, $K$ noise samples yield $K$ renders from a single encoder pass.
We use their spread in the rendered $\Ls\abs$ image as uncertainty,
\begin{equation}
\sigma_{\Ls}=\big(\operatorname{Var}_k \Ls_k\big)^{1/2},\quad
\sigma_{ab}=\big(\operatorname{Var}_k a^{\ast}_k+\operatorname{Var}_k b^{\ast}_k\big)^{1/2},
\label{eq:spread}
\end{equation}
where $\operatorname{Var}_k$ is the unbiased variance over the $K$ samples. There is no variance head
and no calibration target. The maps measure how much the samples disagree. Figures show a single sample per frame. \dawn{} is deterministic and gives no
uncertainty.

\section{Data}
\label{sec:data}

DroneVehicle~\cite{sun2022dronevehicle} provides paired thermal and RGB frames of
$840\times712$ pixels with a 100-pixel white border and annotates five vehicle classes separately in
each modality. We crop the border ($640\times512$) and keep a pair as daytime if the
median linear luminance of its RGB frame is at least 0.20.

\paragraph{Alignment.} Matched box centres provide cross-modal correspondences without feature matching. For box-aligned pairs (3,011), we match
same-class boxes as mutual nearest neighbours and fit a similarity transform with
RANSAC~\cite{fischler1981ransac}, or a homography if the similarity fails only on its residual. A pair
needs at least 6 matches, an inlier ratio of at least 0.6, a median residual of at most
2\,px, a scale change within 5\%, a rotation below $2^\circ$ and a shift below 30\,px, and is then
checked independently by correlating gradient magnitudes. When box annotations are unusable because
labels were copied across modalities or too few vehicles appear, the pair is aligned by a global
translation from the same edge correlation and kept when the correlation has a single clear peak and
the image quadrants agree (2,374 fallback pairs in the final set). Applied to the 3,011 box-aligned pairs, the fallback accepts 1,774, and for these its translation lies within 2\,px (Euclidean) of the box fit's displacement of the frame centre in 78\% of cases and within 3\,px in 93\%. The thermal frame is warped onto the RGB
frame, and each pair carries a mask
\begin{equation}
m=\begin{cases}\texttt{valid}\wedge\neg\,\texttt{moving}, & \text{box-aligned},\\
\texttt{valid}\wedge\neg\,\texttt{vehicles}, & \text{fallback},\end{cases}
\label{eq:mask}
\end{equation}
where \texttt{valid} removes pixels mapped from outside the thermal frame, \texttt{moving} removes
matched vehicles whose residual under the fitted transform exceeds 4\,px, and \texttt{vehicles}
masks all vehicles, dilated by 3\,px and extended by 17.4\,px along their long axis where they may have moved (12\% of a frame on average). Finally, we remove validation and test pairs
whose RGB and thermal perceptual hashes are both within 6 bits of the same training pair (3 pairs).
Tab.~\ref{tab:funnel} summarises the resulting splits. Validation contains 194 box-aligned and 151
fallback pairs, and test 781 and 597.

\begin{table}[t]
\centering
\caption{Number of DroneVehicle pairs after each stage of the data pipeline.}
\label{tab:funnel}
\small
\setlength{\tabcolsep}{3.5pt}
\begin{tabular}{@{}lrrrr@{}}
\toprule
stage & train & val & test & total \\
\midrule
frames in the archives & 17,990 & 1,469 & 8,980 & 28,439 \\
paired and readable & 17,951 & 1,469 & 8,980 & 28,400 \\
daytime (luminance $\ge 0.20$) & 5,818 & 532 & 2,184 & 8,534 \\
aligned (box or edge fit) & 3,662 & 346 & 1,380 & 5,388 \\
\textbf{final} (deduplicated) & \textbf{3,662} & \textbf{345} & \textbf{1,378} & \textbf{5,385} \\
\bottomrule
\end{tabular}
\end{table}

\section{Experiments}
\label{sec:experiments}

\subsection{Setup}
\label{sec:setup}

Both models are trained for 20,000 steps on a single RTX 5090 (32\,GB) with 16 images per step.
\dawnf{} draws 8 timesteps per image, \ie, 128 velocity samples per step, whereas \dawn{} uses a single
pass per image because its input has no noise. The model has 36.8M parameters, of which 8.14M are trainable. We use
AdamW~\cite{loshchilov2019adamw} with a peak learning rate of $2\times10^{-4}$ ($10^{-4}$ for LoRA)
and weight decay 0.01, 1,000 warm-up steps followed by cosine decay~\cite{loshchilov2017sgdr} to zero at step 20,000, gradient
clipping at 1.0, bf16 precision, random flips and random thermal contrast curves. $\Lrender$ uses a random 64 of
the velocity samples of a step, which for \dawn{} means all 16. Every 500 steps we evaluate the EMA
weights (decay $\min(0.999,(1+n)/(10+n))$ after $n$ updates) on 64 box-aligned and 64 fallback
validation pairs and keep the checkpoint with the lowest box-aligned CIEDE2000. Seed 42 of \dawnf{}
took 2.5\,h on the machine it was trained on (87.4 epochs, 17.1\,GB peak memory), and its best
checkpoint is at step 18,000. All other runs of our models, including the three \dawn{} seeds,
use the configuration above and were trained on a second RTX 5090, where \dawn{} takes 0.17\,s per step and
\dawnf{} 0.33\,s. The runs added last, two more seeds of the $\Ls$ correction, four training and
rendering ablations of \dawn{} and one of \dawnf, were trained on a third RTX 5090, mostly three at a
time. Their step times are not comparable and are not reported. At test time \dawnf{} uses NFE\,=\,2 and $K=8$, with noise seeded by the image ID.
We report $\Ls$ PSNR (peak 100), $\Ls$ SSIM~\cite{wang2004ssim} and CIEDE2000~\cite{luo2001ciede2000,sharma2005ciede2000}
within the mask of Eq.~\eqref{eq:mask}, together with full-frame LPIPS (AlexNet)~\cite{zhang2018lpips} and
colourfulness~\cite{hasler2003colourfulness}, all evaluated on a single sample per frame.
For our models the first three are computed on the rendered $\Ls\abs$ image, before the clipping of
Eq.~\eqref{eq:render}, and for the other methods on the $\Ls\abs$ image of their RGB output. Scoring
$\hat y$ instead changes the test scores of \dawn{} (seed 42) and of \dawnf{} (seed 42) by less than
$3\times10^{-4}$.

That initial \dawnf{} (seed 42) run was part of a scripted pipeline in which each stage gates the next. For instance,
the oracle had to reach 24\,dB (it scored 25.08\,dB on all 345 validation pairs) and an overfitting
test 85\% of its oracle (88.7\%). All stages passed, in 3.2\,h. An optional 2,545-parameter lightness
refiner from that pipeline changes the test $\Ls$ PSNR of \dawnf{} by $+0.005$\,dB and is not used
here.

\begin{table}[t]
\centering
\caption{Oracle fit on the validation and test pairs (box: box-aligned, fallb.: fallback), an approximate
ceiling for the representation within the mask of Eq.~\eqref{eq:mask} (optional $\Ls$ correction ablated in
Sec.~\ref{sec:ablations}). $|\Delta\abs|$ is the mean Euclidean chroma error
after upsampling, and edges are the top 10\% of valid pixels by photograph $\Ls$ gradient magnitude.}
\label{tab:oracle}
\small
\setlength{\tabcolsep}{2.4pt}
\begin{tabular}{@{}lccccc@{}}
\toprule
 & \multicolumn{3}{c}{validation} & \multicolumn{2}{c}{test} \\
\cmidrule(lr){2-4}\cmidrule(l){5-6}
 & all & box & fallb. & box & fallb. \\
 & (345) & (194) & (151) & (781) & (597) \\
\midrule
grid only, $\Ls$ PSNR (dB) & 25.08 & 25.13 & 25.02 & 24.45 & 24.81 \\
\quad + generated $\Ls$ correction & 26.85 & 27.04 & 26.60 & 26.49 & 26.29 \\
$|\Delta\abs|$, bilinear & 0.90 & 0.86 & 0.94 & 0.93 & 0.94 \\
$|\Delta\abs|$, BGU & 1.28 & 1.23 & 1.34 & 1.36 & 1.38 \\
$|\Delta\abs|$ at edges, bilinear & 1.45 & 1.39 & 1.51 & 1.46 & 1.52 \\
$|\Delta\abs|$ at edges, BGU & 2.37 & 2.30 & 2.45 & 2.47 & 2.56 \\
CIEDE2000, grid + bilinear & 3.33 & 3.25 & 3.43 & 3.62 & 3.58 \\
\quad + $\Ls$ correction & 2.77 & 2.66 & 2.91 & 2.91 & 3.06 \\
\bottomrule
\end{tabular}
\end{table}

\begin{figure*}[t]
\centering
\footnotesize
\makebox[0.2\textwidth]{Thermal}\makebox[0.2\textwidth]{\dawn}%
\makebox[0.2\textwidth]{\dawnf}\makebox[0.2\textwidth]{pix2pix}%
\makebox[0.2\textwidth]{Photo}\\[1pt]
\includegraphics[width=\textwidth]{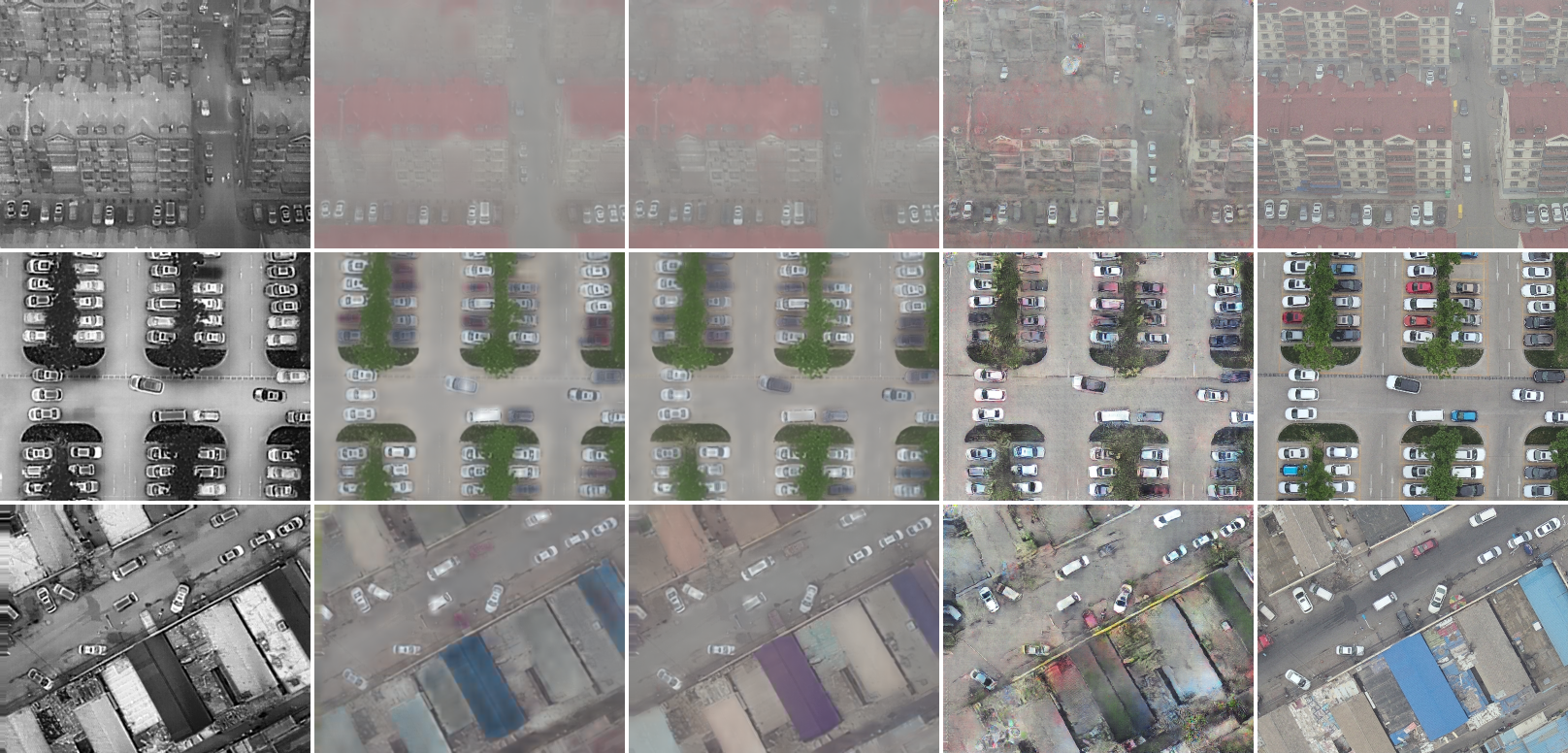}
\caption{Box-aligned test frames at the 10th, 50th and 90th percentile of \dawnf{} (seed 42)
CIEDE2000 (top to bottom). From left to right: thermal input, \dawn{} (seed 42), the scored sample of
\dawnf{} (seed 42), pix2pix, and the photograph. While pix2pix is sharper, it invents
texture and colour absent from the photographs (bottom row), whereas our outputs remain smoother and less
saturated. Images from DroneVehicle~\cite{sun2022dronevehicle}.}
\label{fig:qual}
\end{figure*}

\subsection{Oracle ceiling}
\label{sec:ceiling}

The oracle renders the training targets of each validation pair, \ie, its fitted grid and chroma map,
with the model's code (Tab.~\ref{tab:oracle}). This gives an approximate upper bound for any model with
this output representation. The bound is approximate because the fit is regularised and solved node by
node, and a model could in principle do a little better. On the validation pairs the grid alone reaches
25.08\,dB $\Ls$ PSNR, which means that within each region lightness is largely an affine function of
thermal intensity. On the 781 box-aligned test pairs
it reaches 24.45\,dB and a CIEDE2000 of 3.62, the ceiling we compare test results with. Box-aligned and fallback pairs have similar
ceilings (25.13 vs.\ 25.02\,dB on validation, 24.45 vs.\ 24.81\,dB on test), although the fallback
figures leave out all vehicles (Eq.~\eqref{eq:mask}), and bilinear chroma upsampling is more accurate than BGU-style upsampling~\cite{chen2016bgu} fitted with our node-wise ridge solver from lightness to $\abs$, also
at edges. 

\begin{table*}[t]
\centering
\caption{Comparison on the 781 box-aligned test frames (PSNR and SSIM on $\Ls$). \dawn{} and \dawnf: mean\,$\pm$\,standard
deviation over training seeds 42--44, with \dawnf{}
at NFE\,=\,2 and $K=8$ and one sample scored. Trained on the same pairs, pix2pix is scored from its
8-bit outputs at the final epoch. The global baselines map each pixel's thermal value to a colour
fitted on the training pairs. Oracle fit: the grid and chroma map fitted to the same 781 test
frames (Tab.~\ref{tab:oracle}), an approximate ceiling for the representation (\dag: colourfulness ratio on
the 64 box-aligned validation pairs monitored during training). Colour: colourfulness relative to the
photograph. Vehicles: CIEDE2000 in the vehicle region. Spread: frame-mean $\sigma_{\Ls}$ across the eight
\dawnf{} samples. Time per frame (ms) on the second RTX 5090 (batch size 1, bf16), summed over component
medians for our models and for the generator alone for pix2pix. For \dawnf{} this covers the velocity passes
of all eight samples and one render; timing all eight renders and the spread maps end to end gives 21.19\,ms
(Sec.~\ref{sec:speed}). The global baselines were not timed.}
\label{tab:main}
\footnotesize
\setlength{\tabcolsep}{3.5pt}
\begin{tabular}{@{}lcccccccc@{}}
\toprule
 & PSNR (dB) $\uparrow$ & CIEDE2000 $\downarrow$ & SSIM $\uparrow$ & LPIPS $\downarrow$ & colour $\uparrow$ & vehicles $\downarrow$ & spread $\sigma_{\Ls}$ & ms $\downarrow$ \\
\midrule
affine map & 17.88 & 11.56 & 0.534 & 0.566 & 0.06 & 15.84 & -- & -- \\
tone curve & 18.14 & 11.22 & 0.532 & 0.546 & 0.20 & 15.89 & -- & -- \\
pix2pix~\cite{isola2017pix2pix} & 18.56 & 10.92 & 0.427 & $\mathbf{0.405}$ & $\mathbf{0.89}$ & 15.20 & -- & $\mathbf{1.24}$ \\
\dawnf{} (3 seeds) & $20.01\pm0.05$ & $8.37\pm0.06$ & $0.556\pm0.000$ & $0.535\pm0.001$ & $0.64\pm0.01$ & $12.60\pm0.05$ & $2.20\pm0.03$ & 15.65 \\
\dawn{} (3 seeds) & $\mathbf{20.48\pm0.05}$ & $\mathbf{7.84\pm0.03}$ & $\mathbf{0.559\pm0.000}$ & $0.547\pm0.003$ & $0.68\pm0.01$ & $\mathbf{12.09\pm0.04}$ & -- & 11.02 \\
\midrule
oracle fit (test) & 24.45 & 3.62 & -- & -- & 0.93\dag & -- & -- & -- \\
\bottomrule
\end{tabular}
\end{table*}

\begin{figure*}[t]
\centering
\includegraphics[width=\textwidth]{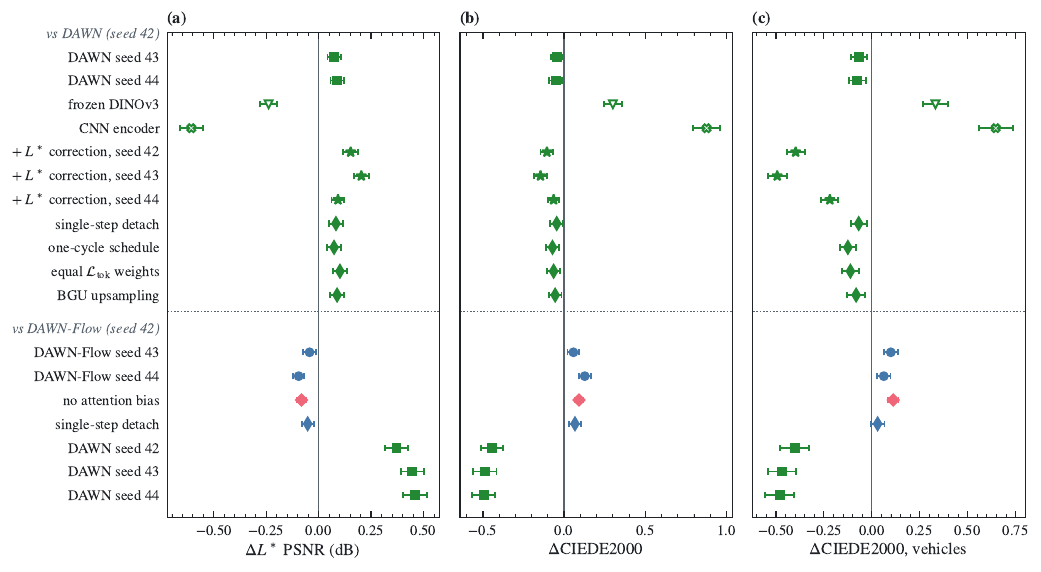}
\caption{Paired differences in (a) $\Ls$ PSNR, (b) CIEDE2000 and (c) CIEDE2000 on vehicles, with 95\%
bootstrap intervals over the 781 box-aligned test frames, relative to \dawn{} with seed 42 (top: two further
seeds, the encoder ablations, three seeds with the $\Ls$ correction and the four training and rendering
ablations of Tab.~\ref{tab:abl}) and to \dawnf{} with seed 42 (bottom: two further seeds, its
attention-bias and single-step-detach ablations, and the three \dawn{} seeds). Intervals reflect test-frame
sampling alone, while the seed rows show retraining variability.}
\label{fig:ablations}
\end{figure*}

\begin{table*}[t]
\centering
\caption{Ablations of \dawn{} on the 781 box-aligned test frames (PSNR and SSIM on $\Ls$), one change per row, seed 42 unless
stated (mean\,$\pm$\,standard deviation over seeds 42--44). Frozen DINOv3: LoRA learning rate 0, which
keeps the adapters at zero, and neither $\Lalign$ nor $\Lcolor$. CNN encoder: a 2.4M-parameter CNN
trained from scratch replaces DINOv3 and LoRA, which also removes the attention bias, the two auxiliary
losses and the detach phase. $+\,\Ls$ correction: 16 more values per token predict a lightness residual
at the resolution of the chroma map. Single-step detach: the detach is lifted at step 3,000 at once
instead of over a 500-step ramp, and the EMA starts without warm-up. One-cycle schedule: a one-cycle
learning-rate and momentum schedule~\cite{smith2019superconvergence} instead of warm-up and cosine decay. Equal $\Ltok$ weights: the brightness and
chroma parts of Eq.~\eqref{eq:flow} weighted equally instead of by their sizes. BGU upsampling:
bilateral guided upsampling~\cite{chen2016bgu} of the chroma map, fitted from $\hat L$ with our ridge solver instead of Chen et al.'s coupled smoothness solve, in training and at
test time. Colour and vehicles as in Tab.~\ref{tab:main}. Time per frame (ms) on the second RTX 5090
(batch size 1, bf16). Rows without a time share the network of \dawn{} or, for the $\Ls$ correction and
BGU, were not timed. Training time per step (s/step) on
the same machine. The runs of the last four rows and seeds 43 and 44 of the correction were trained on a
third RTX 5090, three at a time, and are not timed.}
\label{tab:abl}
\footnotesize
\setlength{\tabcolsep}{3.5pt}
\begin{tabular}{@{}lcccccccc@{}}
\toprule
 & PSNR (dB) $\uparrow$ & CIEDE2000 $\downarrow$ & SSIM $\uparrow$ & LPIPS $\downarrow$ & colour $\uparrow$ & vehicles $\downarrow$ & ms & s/step \\
\midrule
\dawn{} (seed 42) & 20.43 & 7.87 & 0.558 & 0.550 & 0.67 & 12.14 & 11.02 & 0.17 \\
\dawn{} (seeds 42--44) & $20.48\pm0.05$ & $7.84\pm0.03$ & $0.559\pm0.000$ & $0.547\pm0.003$ & $0.68\pm0.01$ & $12.09\pm0.04$ & -- & 0.17 \\
\midrule
\multicolumn{9}{@{}l}{\emph{encoder}} \\
frozen DINOv3 & 20.19 & 8.17 & 0.556 & 0.552 & 0.67 & 12.47 & -- & 0.17 \\
CNN encoder & 19.82 & 8.74 & 0.555 & 0.557 & 0.57 & 12.79 & 3.82 & 0.08 \\
\multicolumn{9}{@{}l}{\emph{output}} \\
$+\,\Ls$ correction (seed 42) & 20.58 & 7.76 & 0.561 & 0.542 & 0.70 & 11.75 & -- & 0.17 \\
$+\,\Ls$ correction (seeds 42--44) & $20.57\pm0.06$ & $7.76\pm0.04$ & $0.560\pm0.001$ & $0.542\pm0.007$ & $0.70\pm0.01$ & $11.77\pm0.14$ & -- & -- \\
\multicolumn{9}{@{}l}{\emph{training and rendering}} \\
single-step detach & 20.51 & 7.82 & 0.559 & 0.548 & 0.69 & 12.07 & -- & -- \\
one-cycle schedule & 20.50 & 7.79 & 0.559 & 0.547 & 0.69 & 12.02 & -- & -- \\
equal $\Ltok$ weights & 20.53 & 7.80 & 0.560 & 0.549 & 0.68 & 12.03 & -- & -- \\
BGU upsampling & 20.51 & 7.81 & 0.559 & 0.542 & 0.69 & 12.06 & -- & -- \\
\bottomrule
\end{tabular}
\end{table*}

\begin{figure*}[t]
\centering
\includegraphics[width=\textwidth]{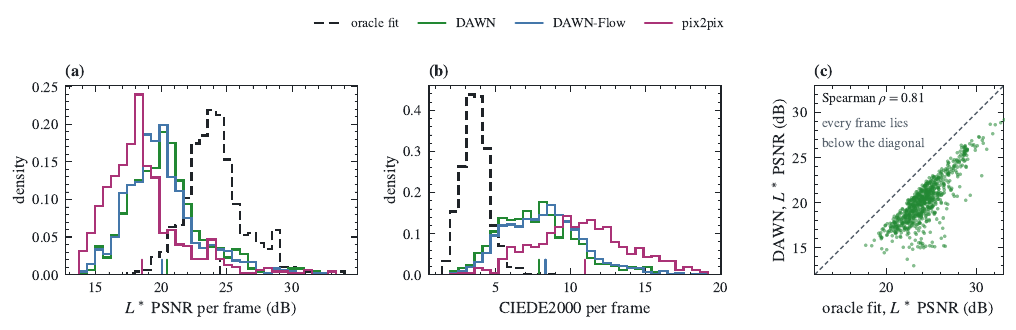}
\caption{Per-frame scores on the 781 box-aligned test frames for the oracle fit (dashed), \dawn{}
(seed 42), \dawnf{} (seed 42) and pix2pix. Step histograms of (a) $\Ls$ PSNR and (b) CIEDE2000, with
vertical ticks along the horizontal axis marking each mean. (c) Per-frame $\Ls$ PSNR of \dawn{} against that
of the oracle fit, with the identity dashed.}
\label{fig:dist}
\end{figure*}

\subsection{Main results}
\label{sec:test}

Tab.~\ref{tab:main} compares \dawn{} with \dawnf, pix2pix~\cite{isola2017pix2pix} and two global
baselines, and Fig.~\ref{fig:qual} shows examples. \dawn{} reaches 20.48\,dB $\Ls$ PSNR and a
CIEDE2000 of 7.84. The grid and chroma map fitted to the same test photographs reach 24.45\,dB and
3.62 (Sec.~\ref{sec:ceiling}). \dawn{} is the most accurate method on every distortion measure, also in the vehicle
regions. Retraining with another seed hardly moves either model. The standard deviation over
seeds is 0.05\,dB for both, while the two models differ by 0.47\,dB. Paired over frames, each
\dawn{} seed is 0.37 to 0.46\,dB more accurate than \dawnf{} (seed 42), and all three
95\% bootstrap intervals exclude zero (Fig.~\ref{fig:ablations}). The advantage is not confined to a
few frames (Fig.~\ref{fig:dist}). Compared frame by frame with \dawnf{} (seed 42) and with pix2pix,
\dawn{} (seed 42) has the higher $\Ls$ PSNR on 76\% and 97\% of the box-aligned test frames and the
lower CIEDE2000 on 74\% and 99\%. The oracle fit scores higher than \dawn{} on every frame, by
3.67\,dB at the median, and the two rank the frames similarly (Spearman $\rho=0.81$). On the 597 fallback test frames \dawn{} scores 20.57\,dB and 7.85, close to its box-aligned scores,
but this comparison favours the fallback frames, whose mask excludes all vehicles, the hardest region
(Eq.~\eqref{eq:mask}). Scored with that same mask, the box-aligned frames reach 21.45\,dB and 7.25 for
seed 42, and the fallback frames 20.49\,dB and 7.89. Fallback frames are about 1\,dB harder, through
their looser alignment, their different content, or both. In the vehicle region (the \texttt{vehicles}
mask, a little larger than the vehicles themselves), the colour error is about 50\% higher than over
the whole frame.

\paragraph{Global baselines.} As lower references, we fit on the training pairs a single affine map
from thermal intensity to $\Ls$ with the mean training chroma, and a tone-and-tint curve that maps each
8-bit thermal value to the mean $\Ls$, $a^\ast$ and $b^\ast$ of the training pixels with that value.
\dawn{} improves on the curve by 2.3\,dB and 3.4 CIEDE2000, yet its LPIPS is not lower. LPIPS barely
separates our models from a global map, whereas it clearly favours pix2pix.

\begin{figure*}[t]
\centering
\includegraphics[width=\textwidth]{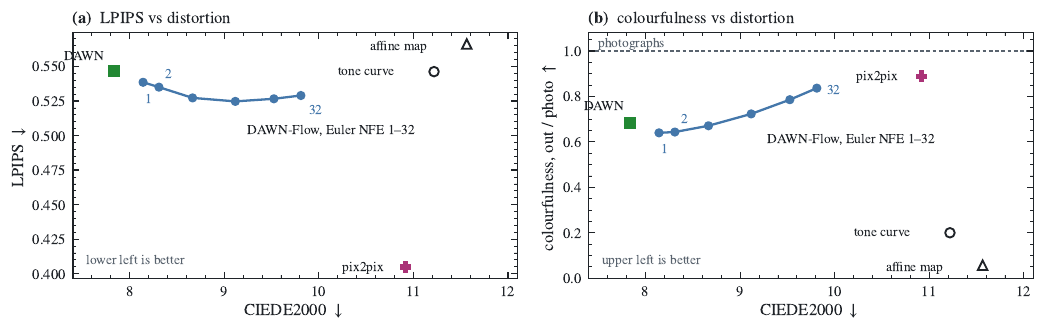}
\caption{Perception against distortion on the 781 box-aligned test frames. (a) LPIPS and (b)
colourfulness relative to the photograph against CIEDE2000 for \dawnf{} (seed 42) across 1 to 32 Euler
steps ($K=8$, one sample scored), \dawn{} (mean and standard deviation over three seeds), pix2pix and the two
global baselines. The dashed line in (b) marks photograph parity. Tab.~\ref{tab:abl} and
Fig.~\ref{fig:ablations} compare the ablations. Increasing NFE moves \dawnf{} toward pix2pix in
colourfulness at the expense of distortion (b), whereas only pix2pix achieves substantially lower LPIPS (a).}
\label{fig:pd}
\end{figure*}

\paragraph{pix2pix.} We train pix2pix with its reference implementation and default settings (an
$L_1$ weight of 100, batch size 1, 200 epochs of which the last 100 decay the learning rate) on the
same 3,662 training pairs, from the single-channel thermal frame at full resolution without resizing
or cropping. Since 640 is not divisible by $2^8$, it uses the reference U-Net with seven downsampling
levels. We test its final epoch in the reference test mode (dropout and batch statistics active),
whereas our checkpoints are selected on validation, and score its 8-bit outputs with the same
evaluation, through which the photographs themselves score a CIEDE2000 and LPIPS of zero. The pix2pix baseline is
the most colourful method and has by far the lowest LPIPS, but it is worse than both of our models on
every distortion measure. Its SSIM is below even that of the global baselines, and its vehicle colours
are barely closer to the photographs than those of the tone curve. Its outputs are sharper and more
textured (Fig.~\ref{fig:qual}), while ours stay close to the conditional mean. The two approaches sit
at opposite ends of the perception--distortion trade-off~\cite{blau2018perception} (Fig.~\ref{fig:pd}).
Unlike \dawnf, pix2pix gives no uncertainty. We did not tune it, and on the GPU it is much faster than any of our
models (Sec.~\ref{sec:speed}).

\begin{figure*}[t]
\centering
\includegraphics[width=\textwidth]{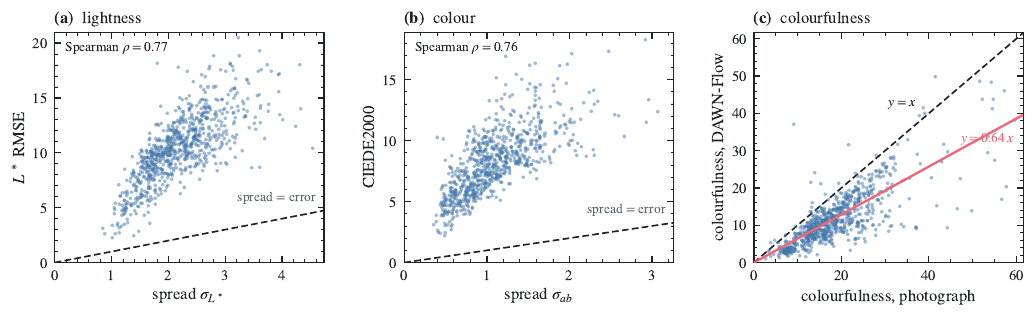}
\caption{Per-frame results of \dawnf{} (seed 42) on the 781 box-aligned test frames (two Euler steps,
$K=8$). (a) $\Ls$ RMSE against frame-mean $\sigma_{\Ls}$ and (b) CIEDE2000 against frame-mean $\sigma_{ab}$,
with the dashed line where error equals spread. (c) Output colourfulness against photograph colourfulness,
with the identity dashed and a solid line of slope 0.64 (the ratio of means).}
\label{fig:perframe}
\end{figure*}

\begin{figure*}[t]
\centering
\includegraphics[width=\textwidth]{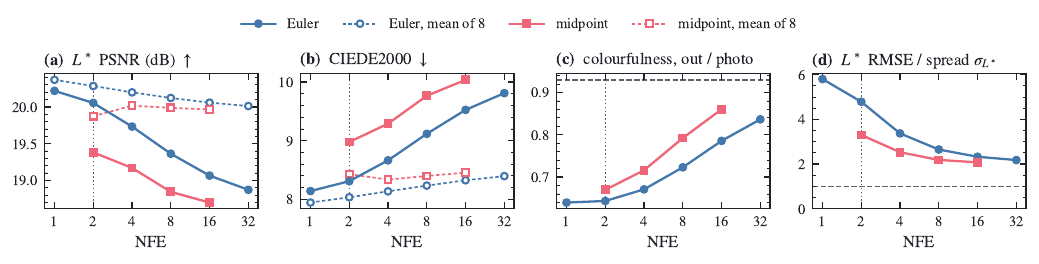}
\caption{Effect of velocity evaluations (NFE) for Euler and midpoint samplers on \dawnf{} (seed 42) over
the 781 box-aligned test frames ($K=8$). (a) $\Ls$ PSNR and (b) CIEDE2000 of a single sample (solid) and of
the eight-sample mean (dashed). (c) Colourfulness relative to the photograph, with the validation oracle ratio
(Tab.~\ref{tab:main}) dashed. (d) Ratio of mean $\Ls$ RMSE to mean $\sigma_{\Ls}$, dashed at unit calibration.
The dotted line marks NFE\,=\,2, used throughout the rest of the paper.}
\label{fig:nfe}
\end{figure*}

\subsection{Ablations}
\label{sec:ablations}

We ablate \dawn{} one change at a time with seed 42, since it trains in half the time of \dawnf{}
(Tab.~\ref{tab:abl}). The intervals below are 95\% paired bootstrap intervals over the 781 test frames
relative to the seed-42 model. Because they cover only the sampling of test frames,
Fig.~\ref{fig:ablations} sets each effect against the differences between seeds. Seed 42 happens to be
the weakest of the three \dawn{} seeds. Seeds 43 and 44 beat it by 0.07 and 0.09\,dB, and both
intervals exclude zero. An ablation has to move \dawn{} further than that before it counts as an effect.

\paragraph{Encoder.} Keeping DINOv3 frozen costs 0.24\,dB [0.20, 0.28] in $\Ls$ PSNR and 0.30 [0.25, 0.35] in CIEDE2000.
The adaptation and its two losses help, although this run cannot separate them. Replacing
DINOv3 with a 2.4M-parameter CNN trained from scratch, mapping the input to the same $32\times40$ token
grid over four stride-2 stages, costs 0.61\,dB [0.55, 0.66] and 0.88 [0.79, 0.96] and lowers
colourfulness from 0.67 to 0.57. It is also the cheapest of our models, at 0.08\,s per training step and
3.82\,ms per frame, its encoder taking 0.84\,ms instead of 7.91. The pretrained features help even a
deterministic model here, although a larger CNN might close part of the gap.

\paragraph{Lightness correction.} The generated $\Ls$ correction of Tab.~\ref{tab:oracle} adds 16
values per token that form a lightness residual at the resolution of the chroma map. Relative to seed
42 it improves every metric, by 0.15\,dB [0.12, 0.19], by 0.11 [0.07, 0.15] in CIEDE2000 and by 0.40
[0.35, 0.44] on vehicles, at the same training cost. Over three seeds the gain is smaller. With the
correction \dawn{} reaches $20.57\pm0.06$\,dB and $7.76\pm0.04$, and without it $20.48\pm0.05$ and
$7.84\pm0.03$. The means differ by 0.09\,dB and 0.07, or 2.2 and 2.6 times their standard errors. It
remains the most accurate configuration in this paper and helps most on vehicles ($11.77\pm0.14$
vs.\ $12.09\pm0.04$).

\paragraph{Training and rendering.} Four further changes move \dawn{} about as far as a change of seed does.
Lifting the detach in a single step without EMA warm-up, a one-cycle learning-rate schedule~\cite{smith2019superconvergence}, equal
weights for the two parts of $\Ltok$ and BGU chroma upsampling change $\Ls$ PSNR by 0.07 to 0.10\,dB and
CIEDE2000 by 0.05 to 0.07 relative to seed 42, while seeds 43 and 44 move it by 0.07 and 0.09\,dB in
PSNR and by 0.04 and 0.05 in CIEDE2000, and each of their intervals overlaps that of a seed
(Fig.~\ref{fig:ablations}). Only BGU lowers LPIPS a little more than another seed does (by 0.008, the
seeds by 0.004 and 0.006). None of these four choices measurably changes the accuracy of \dawn.

\paragraph{Attention bias.} We trained \dawnf{} without DINOv3's attention bias (Eq.~\eqref{eq:gate}
removed, seed 42). It trains in 0.27\,s instead of 0.33\,s per step with 14.8\,GB instead of 17.1\,GB
and scores 19.97\,dB and 8.40, within one standard deviation of the three seeds with the bias.
Zeroing the gates of the trained model costs 0.21 CIEDE2000 (Sec.~\ref{sec:dynamics}), but a model
trained without them does just as well. Within the resolution of three seeds, the attention prior does
not improve accuracy. In \dawn{} the gates stay close to zero. At the selected checkpoints the six
gates of every \dawn{} seed lie between $-0.13$ and 0.06 and those of the three \dawnf{} seeds between
0.25 and 0.99. The regressor relies on the prior far less.

\begin{figure*}[t]
\centering
\includegraphics[width=\textwidth]{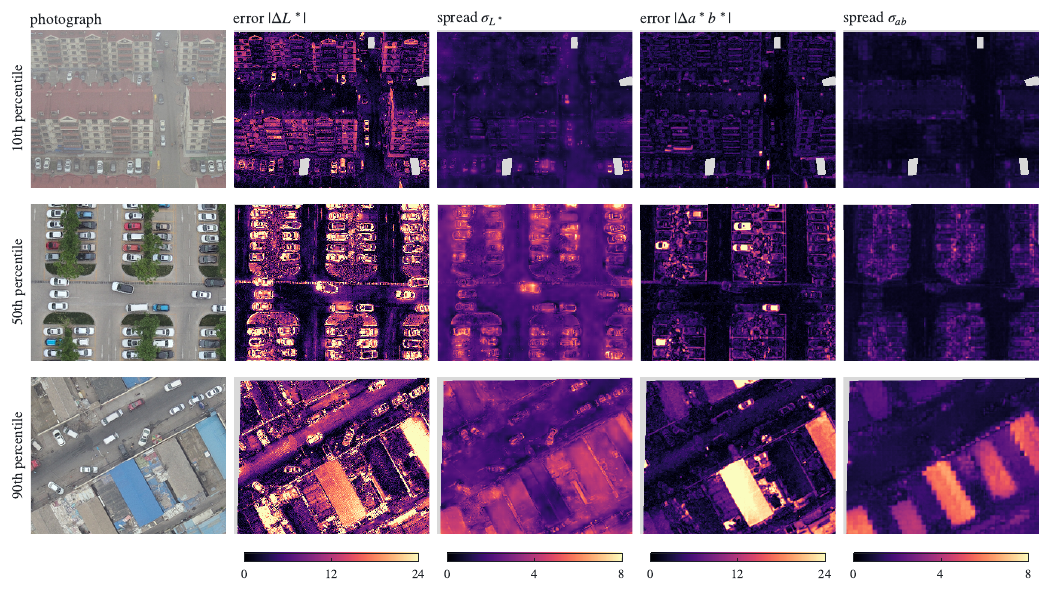}
\caption{Per-pixel error and spread of \dawnf{} (seed 42) on the frames of Fig.~\ref{fig:qual}
(10th, 50th and 90th percentile of its CIEDE2000, top to bottom). From left to right: the photograph,
the absolute $\Ls$ error of the scored sample, the spread $\sigma_{\Ls}$ of the eight samples, the
$\abs$ error and $\sigma_{ab}$. Errors are shown from 0 to 24 and spreads from 0 to 8, a ratio close to
the calibration factors of Sec.~\ref{sec:flow}. Grey pixels lie outside the evaluation mask. Images
from DroneVehicle~\cite{sun2022dronevehicle}.}
\label{fig:errmaps}
\end{figure*}

\subsection{\dawnf: uncertainty and sampling}
\label{sec:flow}

This section analyses \dawnf{} (seed 42), which scores 20.05\,dB and 8.31
with two Euler steps. Fallback frames score 20.11\,dB and 8.37, and the box-aligned frames scored with
the same mask 21.00\,dB and 7.70 (Sec.~\ref{sec:test}). At the same number of network evaluations, the
midpoint method is worse on every metric except colourfulness (19.38\,dB, 8.98, colourfulness 0.67
vs.\ 0.64).

\paragraph{Uncertainty.} Per frame (Fig.~\ref{fig:perframe}), the mean $\sigma_{\Ls}$ and the $\Ls$
RMSE have a Spearman correlation of 0.77, and the mean $\sigma_{ab}$ and CIEDE2000 one of 0.76 (0.77
and 0.67 on fallback frames). The spread is much smaller than the error, however. The mean
$\sigma_{\Ls}$ of 2.18 is about a fifth of the mean RMSE of 10.45. Midpoint sampling ranks frames a
little better ($\rho=0.83$) and underestimates less ($3.3\times$). Per pixel (4,000 random pixels inside the
mask of each frame, 3.1M in all) the ranking is weaker, with a Spearman correlation of 0.33 between
$\sigma_{\Ls}$ and $|\Delta\Ls|$ and 0.57 between $\sigma_{ab}$ and $|\Delta\abs|$. Binned into 20
quantiles of the spread, however, the mean error rises in every bin and stays close to a fixed multiple
of the spread, 3.4 for $\Ls$ and 3.2 for $\abs$. Rescaling the spread by a single factor $a$, as in
post-hoc regression calibration~\cite{levi2022evaluating}, reduces the binned calibration error for
$\Ls$ (the pixel-weighted mean over the bins of $|\overline{|\Delta\Ls|}-a\,\overline{\sigma_{\Ls}}|$)
from 5.61 to 0.60 when $a$ is fitted on the same pixels, and to 0.51 and
0.85 when it is fitted on one half of the frames and applied to the other. Only 20\% of the pixel
errors are smaller than the unscaled $\sigma_{\Ls}$. Fig.~\ref{fig:errmaps} shows the maps on three
frames. On the two harder ones, the chroma spread is large where the chroma error is, on the vehicles
of the middle frame and the blue roofs of the bottom one. The pixels with an $\abs$ error above 12 have
a median $\sigma_{ab}$ at the 85th and 90th percentile of their frame. The lightness spread is a weaker
guide, and on the hardest of the three frames it is unrelated to the lightness error ($\rho=0.03$).

\paragraph{Colour collapse.} At the default setting, several measurements indicate a collapse toward
the conditional mean. The eight samples differ little ($\sigma_{\Ls}=2.18$, and 2.23 with 32 samples),
and their mean scores almost the same as a single sample (20.28\,dB and 8.04 vs.\ 20.05\,dB and 8.31).
The outputs reach 64\% of the photographs' colourfulness, the oracle 93\% and the overfitting run
83\%, and the shortfall is relatively largest for dull scenes (10th percentile 4.08 vs.\ 8.84).

\begin{figure}[t]
\centering
\includegraphics[width=\linewidth]{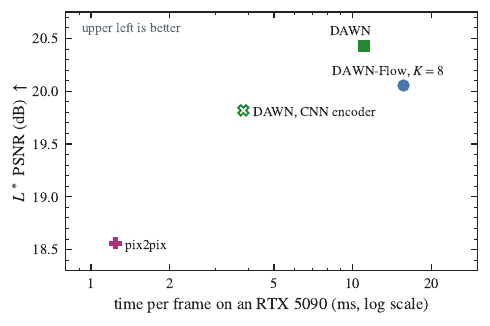}
\caption{$\Ls$ PSNR on the 781 box-aligned test frames against inference time per frame on the second RTX
5090 (batch size 1, bf16, log time axis) for \dawnf{} (NFE\,=\,2, $K=8$), \dawn{} and its CNN-encoder
ablation (all seed 42), and pix2pix. Times sum component medians for our models (one sample rendered for
\dawnf) and give the generator median for pix2pix.}
\label{fig:speedacc}
\end{figure}

\begin{figure*}[t]
\centering
\includegraphics[width=\textwidth]{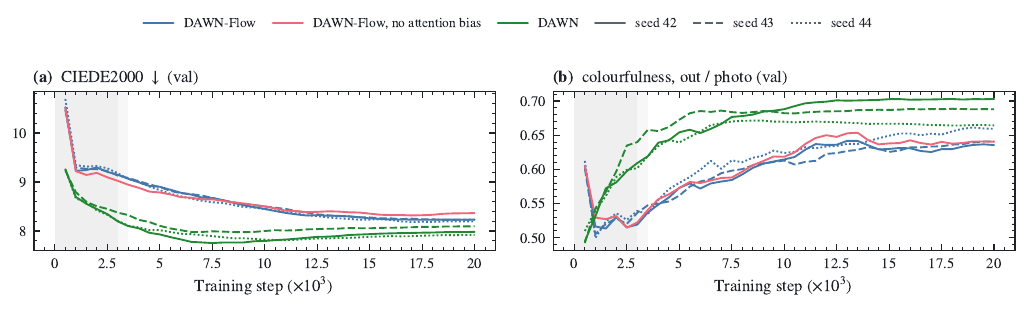}
\caption{Validation trajectories across training for the three seeds of \dawn{} and \dawnf{} and for \dawnf{}
without attention bias. (a) CIEDE2000 and (b) colourfulness ratio on 64 box-aligned validation pairs (EMA
weights, \dawnf{} at NFE\,=\,2). Colour indicates configuration and line style indicates seed. Shaded bands
mark the 3,000-step encoder detach and 500-step gradient ramp.}
\label{fig:runs}
\end{figure*}

\paragraph{Number of steps.} The sampler accounts for most of this collapse (Fig.~\ref{fig:nfe}). Without
reflow~\cite{liu2023rectified}, one Euler step of an exact flow from independent noise returns the conditional mean of the tokens, which is
what \dawn{} is trained to predict. This setting scores best on PSNR and CIEDE2000 (20.22\,dB, 8.14)
with 64\% colourfulness, and two steps barely depart from it. With more steps, the same network
produces more colourful and more varied samples. With 32 Euler steps colourfulness reaches 84\% (86\%
with 16 midpoint evaluations), $\sigma_{\Ls}$ grows to 5.49, the frame ranking improves ($\rho=0.87$)
and the error falls to about twice the spread. Single samples then score worse on distortion metrics
(18.87\,dB, 9.81), as expected when samples move away from the conditional
mean~\cite{blau2018perception}, while LPIPS improves a little (0.525 with 8 steps) and the mean of
eight samples stays between 19.9 and 20.4\,dB~\cite{whang2022deblurring}. The collapse comes mostly from the two-step
sampler, and the number of steps trades distortion for colour and calibrated spread. Per pixel, with 32 steps the error is 1.6 times the spread instead of
3.4 times, and the per-pixel rank correlation rises from 0.33 to 0.37, but the vehicle CIEDE2000 rises
from 12.54 to 14.07, as committing to a specific colour costs more where colour cannot be observed.
Over three seeds, 32 steps give 18.85\,dB and 83\% colourfulness. \dawn{} is more accurate than
\dawnf{} at any number of steps and slightly more colourful than two-step sampling (68\%).
Fig.~\ref{fig:pd} places the main methods on the perception--distortion plane. More steps move \dawnf{}
toward pix2pix in colourfulness, and its distortion grows with them, but only pix2pix reaches a much
lower LPIPS.

\paragraph{Gap to the ceiling.} The overfitting test trains \dawnf{} on 100 frames without augmentation
and evaluates it on 16 of them, where it reaches 23.11\,dB and its oracle 26.06\,dB. Because those 16
frames belong to the training set, this remaining gap cannot come from generalisation. We attribute it to
finite optimisation and the two-step sampler (two Euler steps from noise do not reproduce
an exactly fitted grid), and the rest of the test gap to generalisation,
\begin{equation}
\underbrace{24.45-20.05}_{\approx 4.4\ \text{dB, test}}\;\approx\;
\underbrace{26.06-23.11}_{\approx 3.0\ \text{dB, memorised}}\;+\;
\underbrace{\approx 1.4\ \text{dB}}_{\text{generalisation}} .
\label{eq:gap}
\end{equation}
Because the two gaps are measured in decibels on different frame subsets and schedules, this decomposition
is approximate.

\subsection{Runtime}
\label{sec:speed}

We time the GPU models at batch size 1 in bf16. A component time is the median of 20 timed runs, the
default of our benchmark script, and the time of a model is the sum of its component medians. \dawn{}
needs one encoder pass, one pass of the transformer and one render, 11.02\,ms in all on the second RTX
5090, of which the encoder takes 7.91\,ms. Tab.~\ref{tab:speed} breaks down the cost of \dawnf{} with
$K=8$. On the same GPU its components sum to 15.65\,ms (64\,FPS), and the encoder, run once per frame
regardless of $K$, takes 51\% of that. The render row covers a single sample. Timed as one call with all
eight renders and the spread maps (median of 50 runs), a frame takes 21.19\,ms (47\,FPS), 32\% more
than the sum of its components in the same run. At batch size 1 the encoder and the render are limited
by the host. According to the PyTorch profiler, the GPU computes during 34\% of the encoder's
wall-clock time and 15\% of the render's, but during 87\% of that of the velocity passes. Their cost
depends on the host CPU and software. This is consistent with the encoder running faster on the RTX
5060 Ti than on the RTX 5090 (6.91 vs.\ 7.96\,ms), although the velocity passes take almost four times
as long there. bf16 inference changes the test PSNR and CIEDE2000 of \dawnf{} by less than 0.005 and
none of its metrics in Tab.~\ref{tab:main} by more than 0.02. Its timings therefore hold for the
accuracy we report. We did not repeat this check for \dawn, which has the same architecture.

\begin{figure*}[t]
\centering
\includegraphics[width=\textwidth]{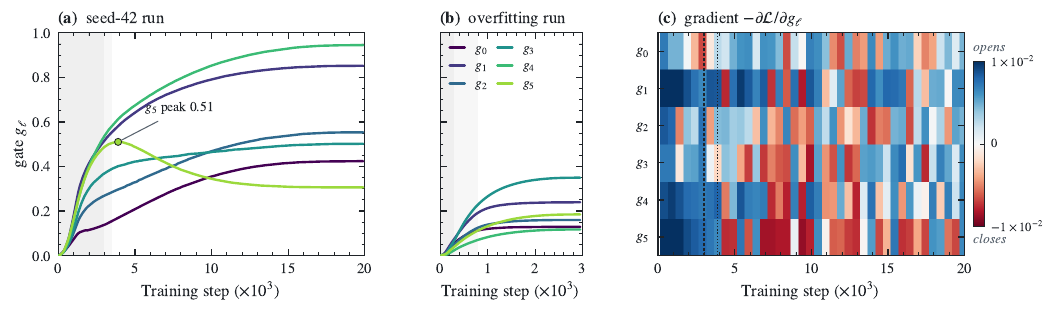}
\caption{Attention gates $g_\ell$ of Eq.~\eqref{eq:gate} in \dawnf. (a) Seed-42 run, logged every 100
steps (live weights), with the peak of $g_5$ marked. The grey band marks the 3,000-step phase in which LoRA
is detached from $\Ltok$ and $\Lrender$, and the paler band next to it marks the 500-step gradient ramp. (b) The same gates in the
100-frame overfitting run (3,000 steps, own schedule). (c) Gate gradient $-\partial\mathcal L/\partial g_\ell$,
one training batch per column on a symmetric-log scale, where blue indicates a gate-opening descent step. The
dashed line marks the end of the detached phase and the dotted line the peak of $g_5$.}
\label{fig:gates}
\end{figure*}

\begin{figure*}[t]
\centering
\includegraphics[width=\textwidth]{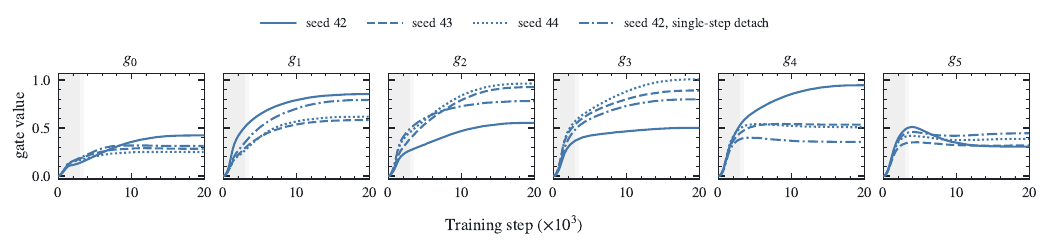}
\caption{The six attention gates $g_\ell$ of \dawnf{} for each training seed and for seed 42 with the
detach lifted in a single step and no EMA warm-up (dash-dot), one panel per block (live weights, logged
every 100 steps). Line style marks the seed as in Fig.~\ref{fig:runs}, and the
grey bands are those of Fig.~\ref{fig:gates}.}
\label{fig:gateseeds}
\end{figure*}

\begin{table}[t]
\centering
\caption{Inference latency of \dawnf{} per $640\times512$ frame in ms (NFE\,=\,2, $K=8$, batch size 1). GPU
columns use bf16 (median of 20 runs) on the second RTX 5090 (Ryzen 7 7800X3D host) and an RTX 5060\,Ti (Core
i7-14700 host). CPU runs on that Core i7-14700 alone in fp32 with eight threads pinned to its performance
cores at its sustained 65\,W power limit; each value is the median over five processes, each timing every
model 50 times across five interleaved rounds after a 150\,s heat-up, with all processes within 7\% of the
median. One render comprises grid slicing, chroma upsampling and CIELAB-to-RGB conversion. End to end times
the encoder, velocity passes, all eight renders and the spread maps in a single call.}
\label{tab:speed}
\small
\setlength{\tabcolsep}{4pt}
\begin{tabular}{@{}lrrrr@{}}
\toprule
component & GMACs & 5090 & 5060\,Ti & CPU \\
\midrule
encoder, once & 52.3 & 7.96 & 6.91 & 200 \\
velocity, 2 NFE $\times$ 8 & 179.7 & 7.01 & 26.31 & 844 \\
render, one sample & 0.06 & 0.68 & 0.76 & 10.9 \\
\midrule
\textbf{total} (sum) & \textbf{232.1} & \textbf{15.65} & \textbf{33.98} & \textbf{1,055} \\
end to end & & 21.19 & 39.94 & 1,153 \\
\bottomrule
\end{tabular}
\end{table}

Fig.~\ref{fig:speedacc} places the timed models by accuracy and time per frame on the second 5090.
The CNN encoder of Sec.~\ref{sec:ablations} cuts \dawn's time to 3.82\,ms but loses accuracy, and
the pix2pix generator (41.8M parameters, 29.9 GMACs) takes 1.24\,ms but is the least accurate of the
four. For a deterministic model on this hardware, the pretrained encoder sets the speed. The same holds
on the Core i7-14700 alone (last column of Tab.~\ref{tab:speed}), where \dawn{} takes 252\,ms per frame,
79\% of it in the encoder, and its CNN-encoder ablation 66\,ms. The pix2pix generator takes 136\,ms there in its
test mode, less than \dawn{} and \dawnf{} (1,153\,ms end to end) but about twice as long as the
CNN-encoder ablation. We have not timed the model on hardware that a drone could carry.

\subsection{Training dynamics and attention gates}
\label{sec:dynamics}

\begin{table}[t]
\centering
\caption{Attention gates $g_\ell$ per block of \dawnf{} (seed 42) and the 100-frame overfitting run (live
weights). ``Final'' is the last logged step (step 19,900,
and step 2,950 for the overfitting run).}
\label{tab:gates}
\small
\setlength{\tabcolsep}{4pt}
\begin{tabular}{@{}lcccccc@{}}
\toprule
block $\ell$ & 0 & 1 & 2 & 3 & 4 & 5 \\
\midrule
step 3,000 (detach lifted) & 0.14 & 0.52 & 0.27 & 0.37 & 0.54 & 0.49 \\
final & 0.42 & 0.85 & 0.55 & 0.50 & 0.94 & 0.31 \\
overfitting run, final & 0.13 & 0.24 & 0.16 & 0.35 & 0.12 & 0.18 \\
\bottomrule
\end{tabular}
\end{table}

Fig.~\ref{fig:runs} shows the validation curves of the three seeds of both models. \dawn{} learns faster than \dawnf. By
step 5,000 its validation CIEDE2000 is below the best that any \dawnf{} seed reaches, and it stays
there. It is lowest between steps 7,500 and 11,000, where its checkpoints are selected, and then rises
a little. \dawnf{} keeps improving almost to the end. During the detached phase of the
seed-42 \dawnf{} run, validation CIEDE2000 dropped from 10.53 at the first validation (step 500) to 9.23 at
step 1,000 and then stayed between 9.15 and 9.30, while patch retrieval (the share of adapted thermal
tokens whose most similar frozen RGB token lies in the $3\times3$ neighbourhood of their own position)
rose from 0.70 to 0.865 at step 3,000 and 0.874 at the end. The held-out training objective was still
at its minimum at the final step, with no sign of memorisation.

Fig.~\ref{fig:gates} and Tab.~\ref{tab:gates} show the six attention gates of Eq.~\eqref{eq:gate} in
\dawnf{} (seed 42). They start at zero and nothing in the objective forces them open, yet all of them
opened, and they did so while the encoder was still detached. At step 3,000, before any velocity
gradient had reached LoRA, five of them were between 0.27 and 0.54. A gate's gradient flows through
the detached bias, which let the generator use DINOv3's attention as a fixed prior before shaping its
features. This is reproducible (Fig.~\ref{fig:gateseeds}). In seeds 43 and 44 the gates also open early,
and over the three seeds five of the six lie between 0.27 and 0.57 at step 3,000. By the end of training, two blocks had opened
almost completely ($g_4=0.94$, $g_1=0.85$). Their attention is then close to the product of their own
and DINOv3's. Every seed ends with two nearly open gates, but which two depends on the seed, $g_2$ and
$g_3$ for seeds 43 and 44 (in seed 44, $g_3$ ends at 1.01, just above 1). Setting gates to zero
at test time, without retraining, shows that the trained network relies mostly on $g_4$. With all six at
zero, CIEDE2000 on the box-aligned test frames rises from 8.31 to 8.52 and colourfulness falls from
64\% to 54\%. Zeroing $g_4$ alone raises CIEDE2000 to 8.47. Zeroing $g_1$ has a smaller effect on
CIEDE2000 (8.34) but lowers $\Ls$ PSNR by 0.09\,dB, more than zeroing any of the other four, each of
which changes CIEDE2000 by at most 0.03. A network trained without gates could compensate, which means
zeroing the gates at test time does not measure what they contribute to training. Trained from scratch
without the attention bias, \dawnf{} scores within the spread of three seeds (Sec.~\ref{sec:ablations}).

The last gate, $g_5$, rose to 0.51 at step 3,900, 400 steps after the end of the ramp, and then
decreased to 0.31. Over the same transition, the median gradient norm reaching the encoder grew from
0.114 in the 1,000 steps before the detach was lifted to 0.185 in the 1,000 steps after the ramp. In
all 8 gradient snapshots up to step 3,900 a descent step would have opened $g_5$, and in 23 of the 32
snapshots after that it would have closed it, compared with 10 to 14 of 32 for the other gates. In every
seed $g_5$ peaks shortly after the ramp (steps 3,800 to 4,200), but it falls by 0.20 only in seed 42 and
by 0.03 in the other two. To test whether the ramp drives this turn, we retrained \dawnf{} with seed 42,
the detach lifted in a single step and no EMA warm-up (Fig.~\ref{fig:gateseeds}, dash-dotted). Its
$g_5$ again peaks at step 3,900, at 0.46, but then falls by only 0.01, and the model scores 20.00\,dB
and 8.38, within the spread of the three seeds. The timing of the peak does not depend on the ramp.
The large fall in seed 42 returns neither with other seeds nor with another schedule, and we
read it as a property of that single run. With the same seed, the schedule also
changes which gates open most ($g_1$ to $g_3$ instead of $g_1$ and $g_4$). In the overfitting run the gates ended between 0.12 and 0.35, in a
different order. With that run's own short schedule this comparison is qualitative, but it is
consistent with the prior being most useful when the model has to generalise across scenes.

\section{Conclusion}
\label{sec:conclusion}

Predicting an affine bilateral lightness grid over thermal intensity and a coarse chroma map from LoRA-tuned
DINOv3 features lets \dawn{} translate daytime thermal drone frames in a single pass. On the 781 box-aligned
DroneVehicle test pairs it reaches 20.48\,dB $\Ls$ PSNR and 7.84 CIEDE2000 (mean of three seeds) in
11.02\,ms per frame on an RTX 5090, against an approximate representation ceiling of 24.45\,dB and 3.62 from
fitting the grid and chroma map directly to the test photographs. Compared with pix2pix trained on the same
pairs, which produces sharper and more colourful outputs at higher distortion, \dawn{} sits at the
low-distortion end of the perception--distortion trade-off. Training the same network by conditional flow
matching (\dawnf) provides per-pixel sample-spread uncertainty whose frame mean ranks frames by error, at a
cost of about half a decibel in $\Ls$ PSNR, while additional sampling steps trade further accuracy for
colourfulness and spread calibration. Ablations show that LoRA-tuned DINOv3 features drive a clear accuracy
gain and a predicted $\Ls$ residual adds a small further improvement ($20.57\pm0.06$\,dB over three seeds),
whereas DINOv3's gated attention prior, the training schedule, the token-loss weighting and guided chroma
upsampling make no measurable difference. Code, training logs and weights for \dawn{} and \dawnf{} are
available at the links below the author names. The weights exclude DINOv3, which carries its own licence, and
no DroneVehicle data are redistributed.

\paragraph{Limitations.} Our evaluation covers a single daytime dataset and sensor setup. Training uses
3,662 DroneVehicle pairs aligned to within a pixel or two with moving vehicles masked rather than registered,
and nighttime scenes, other thermal cameras and embedded drone hardware are not tested. Because we follow
DroneVehicle's official splits after near-duplicate removal, training and test sets can share flight
locations, so test scores reflect generalisation to unseen frames more than to unseen sites. On the baseline
and metric side, pix2pix is evaluated with its default configuration without tuning, we do not benchmark
thermal-specific translation architectures, and the oracle fit is scored on $\Ls$ PSNR, chroma error and
CIEDE2000 rather than SSIM or LPIPS. Finally, only the main models and the $\Ls$ correction (validation
ceiling $25.08\rightarrow26.85$\,dB, trained on \dawn{} alone) use three seeds while the remaining ablations
use one, the bootstrap intervals in Sec.~\ref{sec:ablations} capture test-frame sampling alone, which
attention gates open varies across seeds and schedules, and \dawn{} reuses \dawnf's untuned training schedule
despite peaking much earlier on validation (Sec.~\ref{sec:dynamics}).

\section*{Acknowledgements}
This work began as a technical report on the original \dawn{} code base and has since grown into a full
paper. That code base is available with its complete development history at
\href{https://github.com/NotArsenic/DAWN}{\texttt{github.com/NotArsenic/DAWN}}, and the code released
with this paper is derived from it.

\section*{Use of AI tools}
AI tools were used as a collaborative coding assistant, for debugging and for copy editing. The idea,
the conceptualisation of the method, the design of the experiments, the analysis and the conclusions
are the authors' own.

{\small
\bibliographystyle{ieeenat_fullname}
\bibliography{refs}
}

\end{document}